\documentclass{article} 
\usepackage{iclr2026_conference,times}

\usepackage{amsmath,amsfonts,bm}

\def\eqref#1{equation~\ref{#1}}

\def\1{\bm{1}}

\DeclareMathAlphabet{\mathsfit}{\encodingdefault}{\sfdefault}{m}{sl}
\SetMathAlphabet{\mathsfit}{bold}{\encodingdefault}{\sfdefault}{bx}{n}

\usepackage{hyperref}
\usepackage{url}
\usepackage{amsmath}
\usepackage{graphicx}
\usepackage{todonotes}[small]

\title{Limits of Resolution Equivariance in Fourier Neural Operators}

\author{Alex Colagrande\thanks{corresponding author: alex.colagrande@dauphine.psl.eu} \,$^1$, Paul Caillon$^1$, Eva Feillet$^2$, Alexandre Allauzen$^{1,3}$ \\
$^1$ Miles Team, LAMSADE, Université Paris Dauphine-PSL, Paris, France \\
$^2$ Université Paris-Saclay, CNRS, LISN, France \\
$^3$ ESPCI PSL, Paris, France \\
}

\iclrfinalcopy 
\begin{document}

\maketitle

\begin{abstract}
Fourier Neural Operators are often assumed to generalize across spatial resolutions, enabling training on a coarse grid and deployment on a finer grid. We test this assumption by contrasting two inference-time choices when moving from training resolution $s$ to test resolution $S>s$: running FNO directly at $S$, or running at $s$ and upsampling the prediction to $S$ via Fourier zero-padding. On Darcy flow, we observe that direct fine-grid inference is not reliably beneficial and can be worse than the low-grid-plus-upsampling baseline. We further analyze layerwise spectra and find that, under Fourier truncation, intermediate representations increasingly concentrate energy in low frequencies, with high-frequency output produced mainly by late nonlinear/decoder stages. This offers a mechanistic explanation for why FNO can perform well while retaining few modes, yet remain sensitive under resolution shifts. Our findings highlight a simple but strong baseline for cross-resolution evaluation and point to nonlinear aliasing as a key obstacle to zero-shot resolution equivariance. 
\end{abstract}

\section{Introduction and Related Works}
Fourier Neural Operators (FNOs) are a widely used approach for learning solution operators of parametric PDEs from data~\citep{FNO}. A common working assumption is that, because FNO performs global convolutions in Fourier space with a fixed number of retained modes, one can train on a coarse grid and deploy the same model on a finer grid with little or no degradation. This intuition is appealing when high-resolution simulations are costly, yet high-resolution predictions are desired.
We refer to the works of~\cite{kovachki2023neural, Survey_NO_new, practical_FNO} for a broader perspective on operator learning and principled extensions of neural architectures to function spaces.
\newline
Recent work, however, suggests that resolution-agnostic parameterizations do not automatically imply resolution-equivariant inference. \citet{gao_disc_inv} formalize discretization mismatch errors in neural operators and show that resolution bias can persist (and accumulate through depth) under shifts in discretization. Complementarily,~\citet{sakarvadia_disc_inv} show that ``zero-shot super-resolution'' blurs interpolation across grids with extrapolation to unseen high-frequency content, and report systematic failures driven by aliasing and discretization mismatch.
Motivated by these findings, we study cross-resolution behavior using (i) a controlled upsampling baseline and (ii) layerwise spectral measurements, which together help disentangle the effect of nonlinear aliasing from the linear spectral convolution. While aliasing does not provide reliable cross-resolution generalization, our spectral analysis suggests it may contribute to how FNOs remain effective even when retaining only a small number of Fourier modes.
\newline
Concretely, we study zero-shot deployment from a training resolution $s$ to a finer resolution $S>s$ by comparing two inference protocols: running the trained FNO directly at $S$, or running it at $s$ and upsampling the \emph{prediction} to $S$ via Fourier zero-padding (band-limited interpolation). This isolates the effect of executing the network at an unseen discretization. On Darcy flow, direct fine-grid inference is not reliably beneficial and can underperform the low-grid-plus-upsampling baseline. 
We track layerwise spectra at fixed resolution and observe that, under Fourier truncation, activations increasingly concentrate energy in low frequencies, while high-frequency output is produced mainly by late nonlinear/decoder stages. 
Aliasing produced by non-linear activations thus allows the retention of high-frequency modes even when training at low resolutions, but can also be detrimental to high-resolution inference.
Overall, we provide a simple baseline for cross-resolution evaluation and a spectral diagnosis that helps reconcile why FNOs can work well with few modes yet remain fragile under resolution shifts.

\section{Methodology}

\textbf{Zero-shot cross-resolution setting.}
We study zero-shot deployment from a training resolution $s$ to a finer resolution $S>s$. We train an FNO model $\mathcal{M}$ on a dataset $\mathcal{D}_s$ sampled on an $s\times s$ grid, and evaluate it without retraining on a test set $\mathcal{D}_S$ sampled on an $S\times S$ grid. This setting matches the common ``train coarse, deploy fine'' motivation in operator learning~\citep{FNO,Survey_NO_new}. 
Unless stated otherwise, each spectral convolution retains a number of Fourier modes $K$ (per spatial dimension), as in~\citet{FNO}. When training at low resolution, all the modes are kept.

\textbf{Inference protocols.}
Given an input $x_S$ at resolution $S$, we compare:
(i) \emph{Direct-$S$ inference}: evaluate $\mathcal{M}$ directly on $x_S$ to obtain $\hat y_S$; and
(ii) \emph{Low+pad baseline}: downsample $x_S$ to resolution $s$ by Fourier truncation, run inference at the training resolution, and upsample the \emph{prediction} back to $S$ by Fourier zero-padding.
Formally, letting $T_{S\to s}$ denote Fourier truncation and $U_{s\to S}$ denote Fourier zero-padding, the baseline is
\begin{equation}
\hat y_{s\to S} \;=\; U_{s\to S}\!\left(\mathcal{M}\!\left(T_{S\to s}(x_S)\right)\right).
\end{equation}
This baseline cannot recover frequencies outside the training band, but it avoids executing the network at an unseen discretization and provides a strong reference for cross-resolution evaluation.

\textbf{Layerwise spectral diagnostics.}
To localize where frequency content is created or suppressed, we record intermediate activations at a fixed resolution and compute their radially averaged power spectra across layers~\citep{dieleman2024spectral}. We track how spectral energy redistributes relative to the retained band induced by the spectral convolutions. 
These measurements test whether cross-resolution degradation correlates with spectral leakage and aliasing induced by pointwise nonlinearities and decoder stages when the model is executed at a resolution different from training.

\textbf{Dataset.}
We evaluate on the Darcy Flow benchmark from~\citet{FNO}, which models flow through porous media:
\begin{equation}
    \begin{aligned}
        \nabla \cdot (a(x)\nabla u(x)) &= f(x), \quad &&x \in (0,1)^2, \\
        u(x) &= 0, \quad &&x \in \partial(0,1)^2,
    \end{aligned}
\end{equation}
where $a(x)$ is the permeability coefficient, $f(x)$ is the forcing term and $u(x)$ is the solution. The learning task is to approximate the operator $a \mapsto u$. We follow the standard protocol with $1000$ training samples and $200$ test samples. The released data are available at $421\times 421$, enabling evaluation across a wide range of discretizations.

\textbf{Model.}
We adopt the original FNO architecture and official implementation.\footnote{\label{note1}\url{https://github.com/ixScience/fourier_neural_operator}~\citep{FNO}}
The network comprises a lifting MLP, followed by $L$ FNO blocks; each block sums (i) a spectral convolution applied to a truncated set of Fourier modes and (ii) a residual branch given by a spatially shared linear map, followed by a GeLU nonlinearity. A final two-layer MLP projects latent features to the output.

\textbf{Cross-resolution protocol.}
We train at resolution $s=85$ and evaluate zero-shot at higher resolutions $S>s$. Unless stated otherwise, we reuse the normalizer fitted on the training set at resolution $s$ for all test resolutions. For band-limited resampling (used in our baseline and in controlled comparisons), we downsample via Fourier truncation and upsample via Fourier \emph{zero-padding} of modes.

\textbf{Spatial padding.}
Since Darcy flow is non-periodic, spectral convolutions in FNO are commonly implemented with \emph{spatial} padding in the physical domain (standard setting: padding $=10$ in~\citet{FNO}). In our cross-resolution experiments, spatial padding can substantially affect performance; we therefore report results both without spatial padding and with padding $=10$.

\section{Results and Analysis}

As noted by~\citet{FNO}, FNO tends to generalize across resolutions when the fine-grid data is (nearly) band-limited to the training Nyquist frequency (the relevant cutoff in the interpolation vs. extrapolation framing of~\citet{sakarvadia_disc_inv}). In that case, the new content at resolution $S$ can be quantified by the relative energy in modes beyond the training band. We therefore focus on a controlled best-case setting where the additional high-frequency modes at $S$ are set to zero via Fourier padding, and complement this evaluation with a layerwise spectral analysis to track how frequency content is created, suppressed, or redistributed under truncation. We note that executing the network on an unseen grid can still induce nonlinear aliasing that folds unresolved frequencies back into the learned band.
Here, we use \emph{aliasing} in the signal-processing sense: on a discrete grid, frequency content generated beyond the Nyquist limit by non-linear functions folds back into lower frequencies. Our results below show that this folding is harmful for zero-shot execution at an unseen resolution, yet it can also facilitate cross-mode communication when the spectral convolution retains only a limited bandwidth.

\begin{figure}[t]
\centering
\begin{minipage}{0.19\textwidth}
\includegraphics[width=\linewidth]{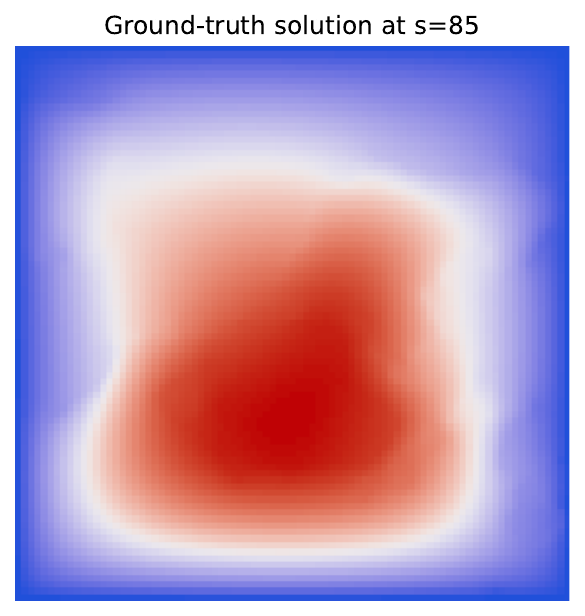}
\end{minipage}\hfill
\begin{minipage}{0.19\textwidth}
\includegraphics[width=\linewidth]{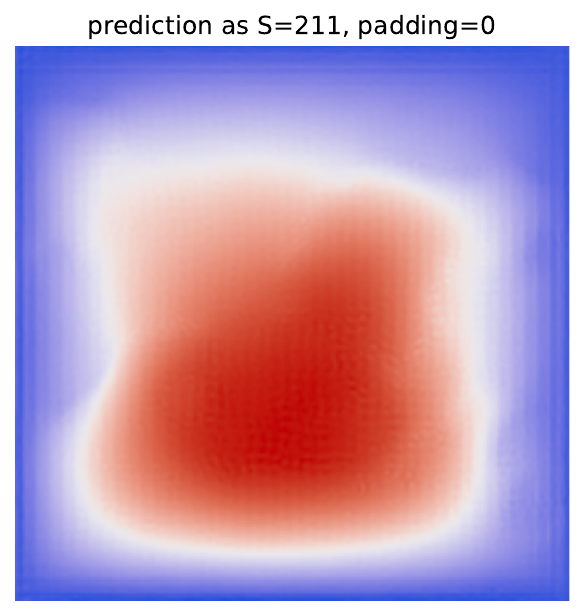}
\end{minipage}\hfill
\begin{minipage}{0.19\textwidth}
\includegraphics[width=\linewidth]{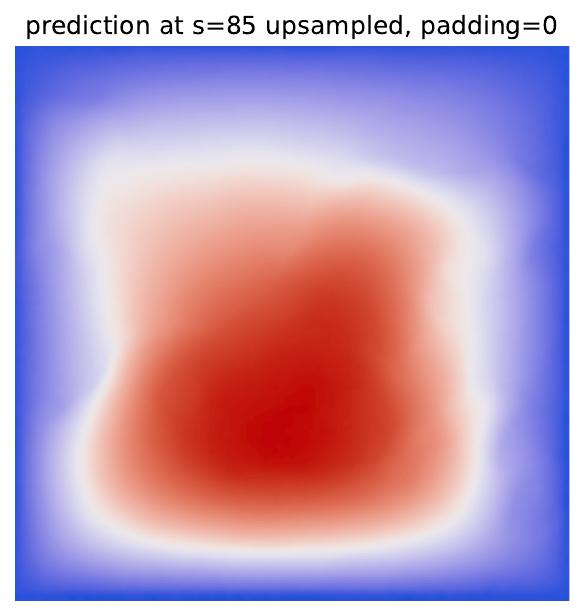}
\end{minipage}\hfill
\begin{minipage}{0.19\textwidth}
\includegraphics[width=\linewidth]{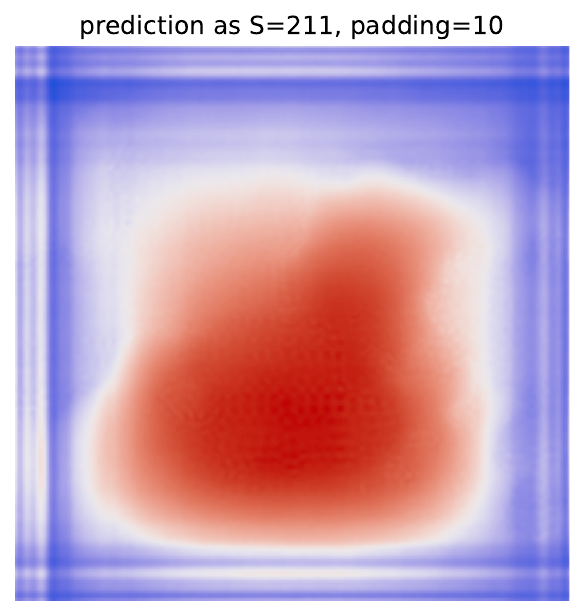}
\end{minipage}\hfill
\begin{minipage}{0.19\textwidth}
\includegraphics[width=\linewidth]{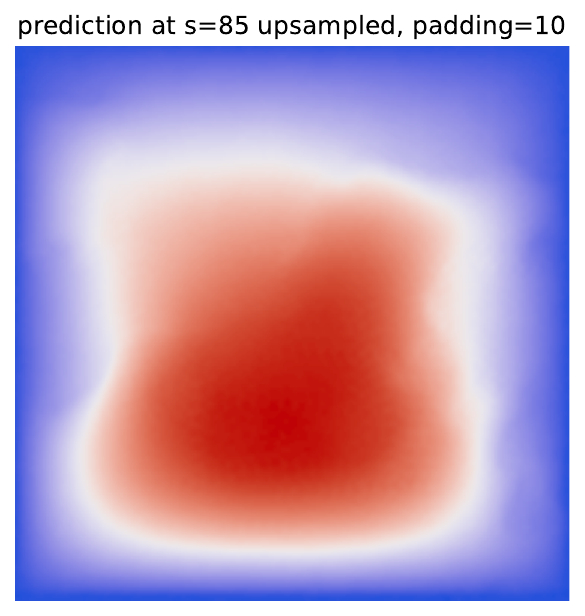}
\end{minipage}

\par\medskip

\begin{minipage}{0.19\textwidth}
\includegraphics[width=\linewidth]{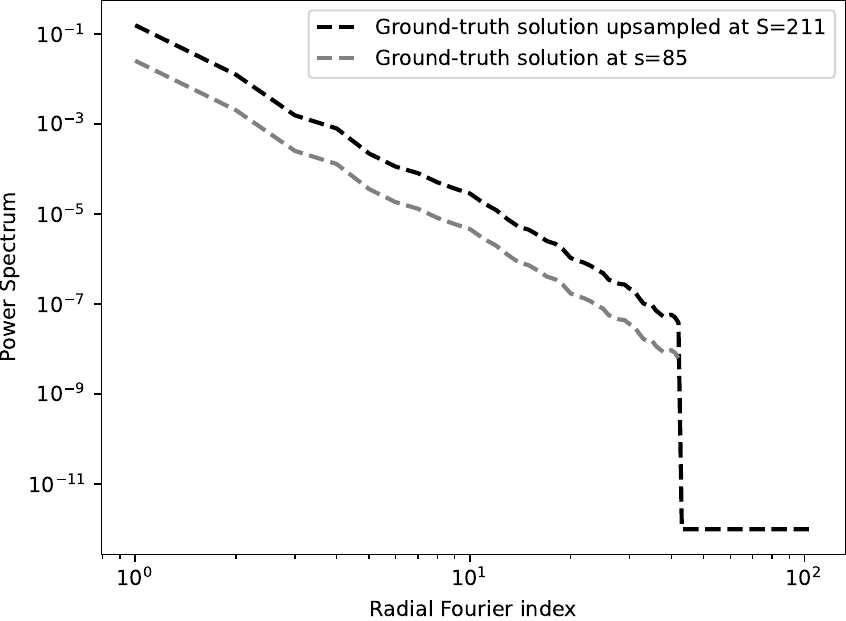}
\end{minipage}
\begin{minipage}{0.19\textwidth}
\includegraphics[width=\linewidth]{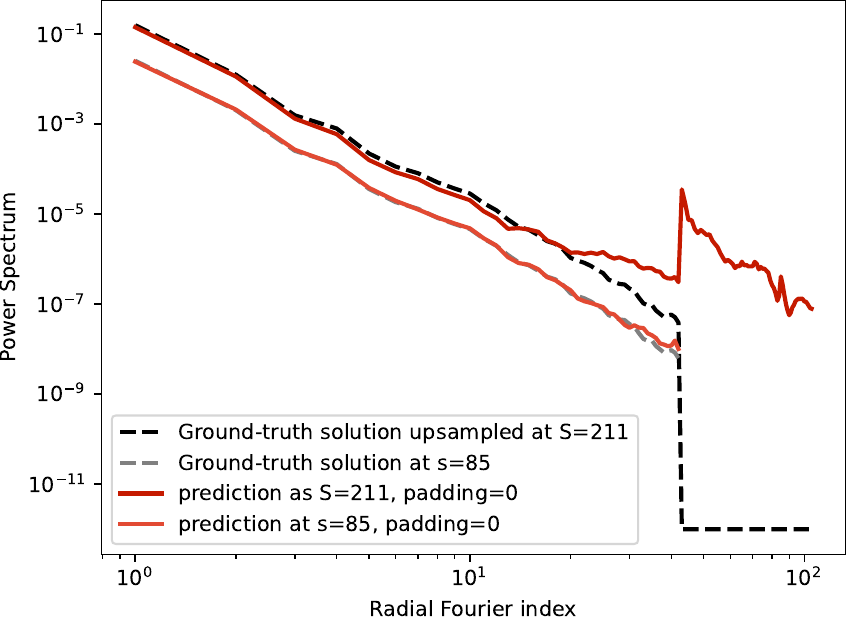}
\end{minipage}
\begin{minipage}{0.19\textwidth}
\includegraphics[width=\linewidth]{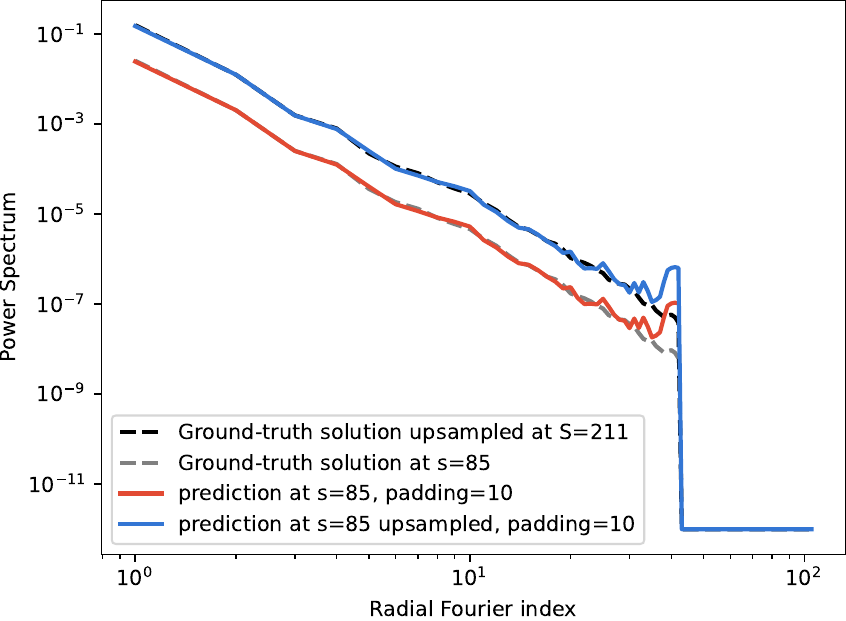}
\end{minipage}
\begin{minipage}{0.19\textwidth}
\includegraphics[width=\linewidth]{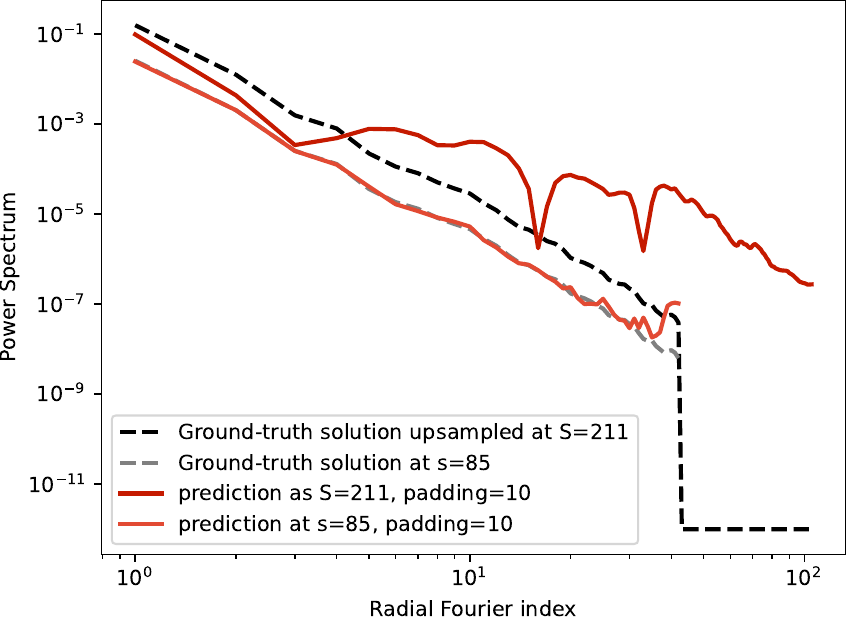}
\end{minipage}
\begin{minipage}{0.19\textwidth}
\includegraphics[width=\linewidth]{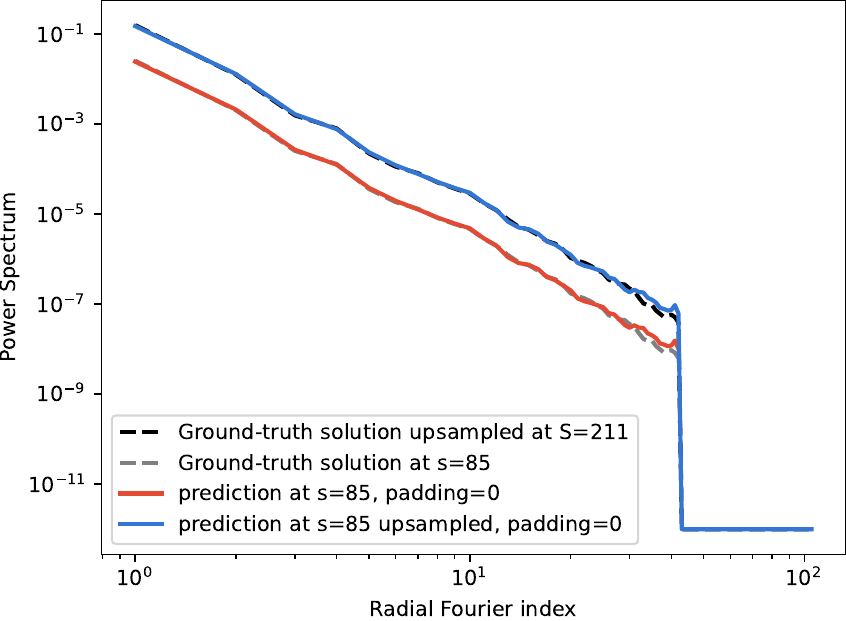}
\end{minipage}

\caption{
\textbf{Cross-resolution inference vs.\ a band-limited upsampling baseline ($s=85 \rightarrow S=211$).}
\textbf{Top:} ground-truth at $s=85$, Direct-$S$ prediction at $S=211$, and Low+pad prediction (run at $s$ then upsample the output to $S$ via Fourier zero-padding), shown with \emph{no spatial padding} and \emph{spatial padding $=10$}.
\textbf{Bottom:} corresponding radially averaged power spectra.
Direct evaluation at $S$ produces a spurious high-frequency tail beyond the training Nyquist frequency, whereas Low+pad preserves the spectral decay of the band-limited target. Spatial padding reduces boundary artifacts but does not recover missing high-frequency content.
}
\label{fig:five-in-row}
\end{figure}

\textbf{Direct fine-grid execution can be worse than Low+pad.}
We compare two inference protocols for $s=85 \rightarrow S=211$: (i) \emph{Direct-$S$} (evaluate the trained model on the $S\times S$ grid) and (ii) \emph{Low+pad} (downsample input to $s$, evaluate at $s$, then upsample the \emph{prediction} to $S$ by Fourier zero-padding). Figure~\ref{fig:five-in-row} shows that Direct-$S$ is not reliably beneficial and can underperform Low+pad.

The spectra explain the gap: Direct-$S$ produces a spurious high-frequency tail and perturbs the low-frequency band, consistent with aliasing induced by nonlinearities when the model is executed at an unseen discretization. Spatial padding mitigates boundary artifacts but does not remove the spurious spectral tail.
Because the linear parts of FNO (spectral convolution and residual linear map) are mode-wise, these resolution-dependent spectral corruption points to the nonlinearities (and their discretization) as the primary source of the mismatch.

\begin{figure}[t]
\centering
\begin{minipage}{0.25\textwidth}
\includegraphics[width=\linewidth]{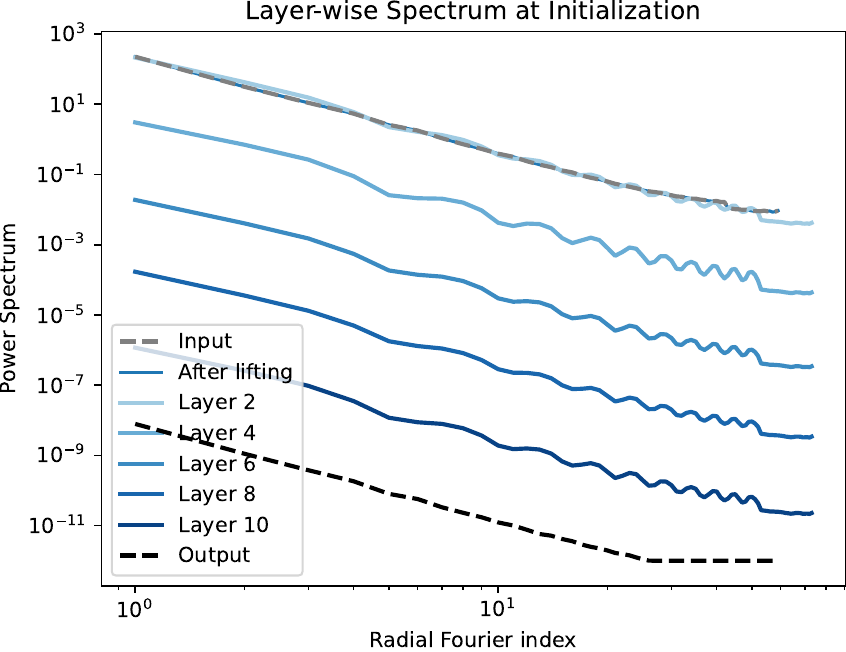}
\end{minipage}\hfill
\begin{minipage}{0.25\textwidth}
\includegraphics[width=\linewidth]{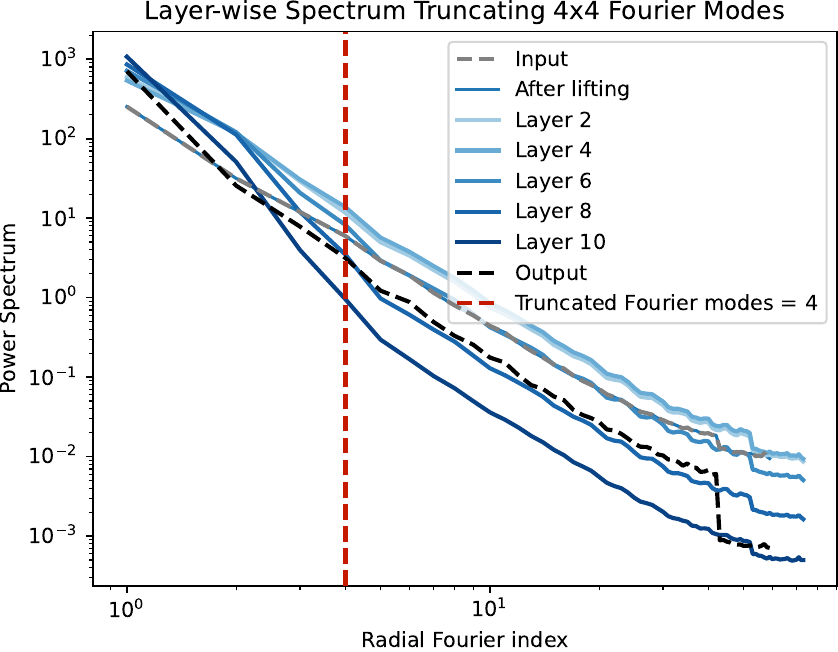}
\end{minipage}\hfill
\begin{minipage}{0.25\textwidth}
\includegraphics[width=\linewidth]{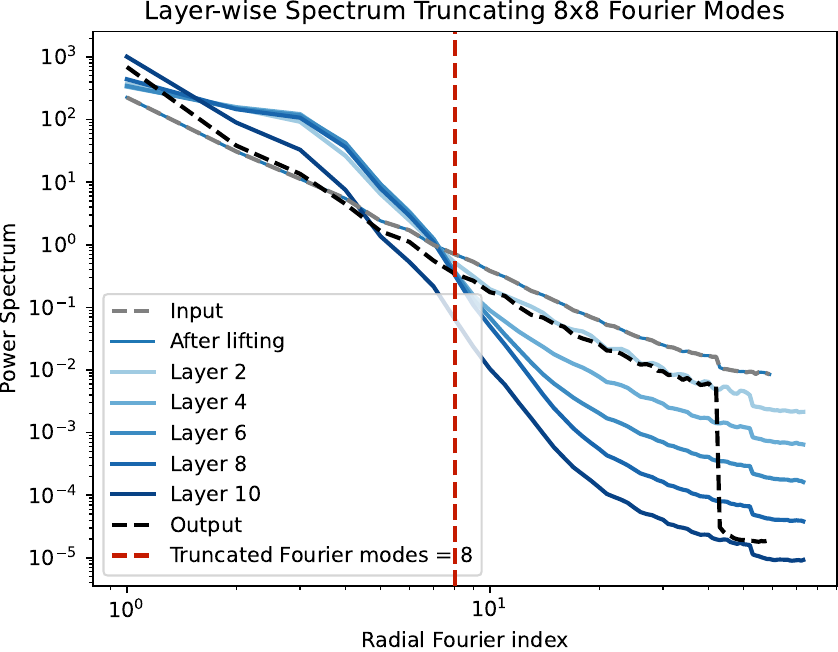}
\end{minipage}\hfill
\begin{minipage}{0.25\textwidth}
\includegraphics[width=\linewidth]{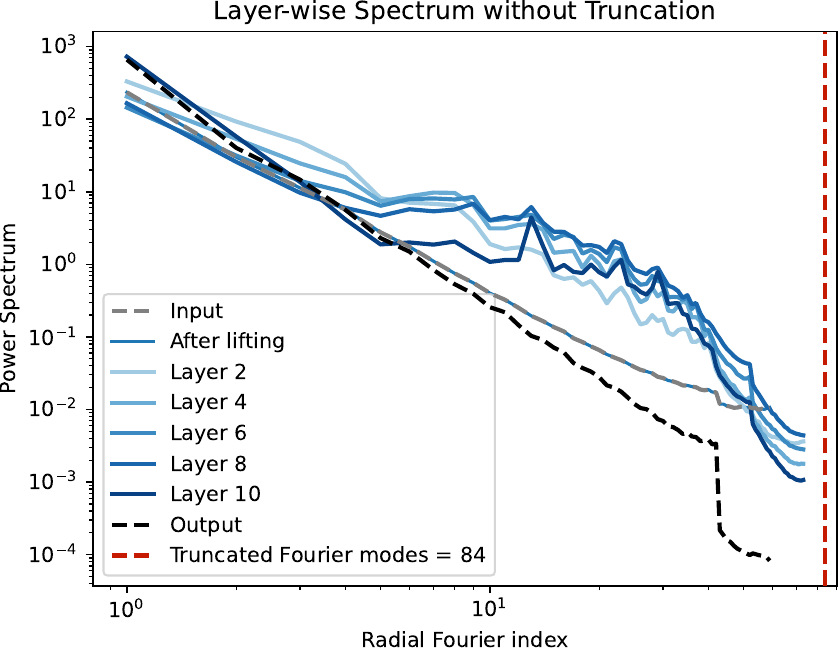}
\end{minipage}\hfill
\caption{
\textbf{Layerwise radially averaged power spectra under different Fourier truncation levels.}
From left to right: initialization (no truncation), then truncation to $4\times4$, $8\times8$, and $64\times64$ modes.
Curves show the radially averaged power spectrum of the input, intermediate feature maps (after lifting and successive layers), and final output. 
The vertical red dashed line indicates the maximum retained Fourier mode (truncation threshold). 
As depth increases, energy is progressively reorganized toward lower radial frequencies, while frequencies beyond the truncation boundary sharply decay. 
At initialization, and also when all Fourier modes are retained, we do not observe any concentration of feature-map energy in the lowest frequencies.
}
\label{fig:layer_wise_spectrum}
\end{figure}

\textbf{Layerwise spectra suggest an encode-process-decode pattern under truncation.}
Figure~\ref{fig:layer_wise_spectrum} tracks spectra of intermediate feature maps for different truncation levels. Under severe truncation (e.g.\ $4\times4$ and $8\times8$ modes), intermediate activations progressively concentrate energy in the retained low-frequency band across depth, while the final output recovers substantially more high-frequency energy than the last activation. This supports an \emph{encode-process-decode} interpretation: the network processes a low-band latent representation (spread across channels), and the projection head reconstructs higher frequencies from it.
Notably, the last activation is more strongly band-limited than the final output, suggesting that high-frequency content is produced mainly by the decoder from low-band latent features rather than being propagated explicitly through retained Fourier modes.

\textbf{Aliasing from nonlinearities: a double-edged sword.}
Within a standard FNO block, the spectral convolution and residual branch are linear and act mode-wise on the retained coefficients. Pointwise nonlinearities are the main source of cross-mode interactions and can induce spectral leakage. On a discrete grid, this interaction manifests as \emph{aliasing}: frequency content generated beyond the Nyquist limit folds back into lower frequencies. 
Compressing information from a large number of Fourier modes into a significantly smaller subset is, in general, infeasible at the intrinsic dimensionality of the problem. However, lifting the representation to a higher-dimensional embedding space can provide sufficient capacity to reorganize and redistribute this information, enabling an effective low-frequency encoding of high-frequency content. 
This provides a plausible mechanism
for the learned spectral reorganization in Figure~\ref{fig:layer_wise_spectrum}, but it also explains why executing the network at
an unseen resolution can be harmful: grid-dependent leakage/aliasing can corrupt the learned band,
consistent with the spurious high-frequency tail observed under Direct-S inference in Figure~\ref{fig:five-in-row}.

\section{Discussion and Conclusion}
The original FNO work~\citep{FNO} reported encouraging zero-shot cross-resolution behavior, but more recent studies~\citep{gao_disc_inv,sakarvadia_disc_inv} have questioned the robustness of such generalization and highlighted discretization- and aliasing-related failure modes. Our results support this view: when high-frequency modes are absent from low-resolution observations, they are not identifiable in general, so zero-shot recovery of the correct fine-scale content cannot be expected without additional assumptions. Even in a controlled best-case regime (band-limited resampling), we find that evaluating FNO directly at a higher resolution can underperform a simple practitioner baseline: predict at the training resolution and upsample the \emph{prediction} via Fourier zero-padding.

At the same time, our layerwise spectral measurements help reconcile why FNO can still perform well under strong spectral truncation. We observe an encode--process--decode pattern in which intermediate feature maps become increasingly low-band with depth, while the output recovers higher-frequency content primarily in late stages. This points to a dual role of nonlinearities: under resolution shifts, nonlinear aliasing can corrupt the learned band and degrade cross-resolution inference, while in-distribution nonlinear mixing provides a plausible pathway for cross-mode communication that enables strong performance even when only a small number of Fourier modes is retained. This mechanism allows FNO to exploit high-dimensional feature representations while operating with limited spectral bandwidth.

This suggests that FNO-like models may be particularly well suited for few-shot super-resolution settings, where additional high-resolution information is available and accurate reconstruction is information-theoretically feasible. We consider this direction as promising for future work. Future work could combine limited multi-resolution supervision with anti-aliased and nonlinearity-aware designs to reduce grid-dependent spectral folding at inference time. This could help to improve robustness across discretizations while preserving the efficiency benefits of spectral truncation.

\subsubsection*{Acknowledgments}

This work was granted access to the HPC resources of IDRIS under the allocation A0191016927 made by GENCI. This work has received support from the French government, managed by the National Research Agency, under the France 2030 program with the reference “PR[AI]RIE-PSAI” (ANR-23-IACL-0008) and "PEPR-SHARP" (ANR-23-PEIA-0008).

\bibliography{iclr2026_conference}
\bibliographystyle{iclr2026_conference}

\appendix

\section{Experimental Details}
\subsection{Darcy Flow}
\label{app:darcy}
This benchmark represents the flow through porous media. 2D Darcy flow over a unit square
is given as follows:
\begin{equation}\label{eq:darcy}
    \begin{aligned}
        \nabla \cdot (a(x)\nabla u(x)) &= f(x), \quad x \in (0,1)^2 \\
        u(x) &= 0, \quad x \in \partial (0,1)^2
    \end{aligned}
\end{equation}
In this dataset, the input is represented by the permeability coefficient $a$, and the corresponding output is the solution $u$. The permeability is sampled as $a \sim \mu$, where $\mu = \psi_{\#}\mathcal{N}\!\left(0,(-\Delta+9I)^{-2}\right)$ and $\Delta$ is equipped with zero Neumann boundary conditions. The nonlinearity $\psi:\mathbb{R}\to\mathbb{R}$ is applied pointwise and maps positive values to $12$ and negative values to $3$. The forcing term is fixed to $f(x)=1$. 

Equation ~\ref{eq:darcy} is modified in the form of a temporal evolution as follows:
\begin{equation}
    \partial_t u(x,t) - \nabla \cdot (a(x) \nabla u(x,t)) = f(x), \quad x \in (0,1)^2, \label{eq:temporal}
\end{equation}
This construction is a prototypical model for heterogeneous coefficients arising in applications such as permeability in subsurface flows and material microstructures in elasticity. Solutions $u$ are computed with a second-order finite-difference scheme on a $421\times 421$ grid, and lower resolutions are obtained by downsampling this reference. For training, 1000 samples are used, 200 samples are generated for testing, and different cases contain different medium structures.

\subsection{Training Details}
All models were trained following standard procedures. Optimization was performed using AdamW with an initial learning rate of $10^{-3}$, scheduled via CosineAnnealingLR down to a final learning rate of $10^{-6}$.

The performance achieved in our experiments is consistent with previously reported results in the literature, as we adopt the same training codebase and closely match common hyperparameter settings.
\end{document}